\pdfoutput=1

\documentclass[11pt]{article}

\usepackage[final]{acl}

\usepackage{times}
\usepackage{latexsym}

\usepackage[T1]{fontenc}

\usepackage[utf8]{inputenc}

\usepackage{array}
\usepackage{multirow}
\usepackage{ragged2e}
\usepackage{verbatim}
\usepackage{amsmath}
\usepackage{amssymb}   
\usepackage{amsfonts}  

\usepackage{microtype}

\usepackage{inconsolata}

\usepackage{graphicx}

\title{Zero-shot Large Language Models for Automatic Readability Assessment}

\author{Riley Grossman \\
  New Jersey Institute of Technology\\
  Newark, NJ \\
  \texttt{rag24@njit.edu} \\\And
  Yi Chen \\
  New Jersey Institute of Technology\\
  Newark, NJ \\
  \texttt{yi.chen@njit.edu} \\}

\newcommand{\yc}[1]{\textcolor{green}{#1}}
\newcommand{\wg}[1]{\textcolor{blue}{#1}}
\renewcommand{\yc}[1]{}
\renewcommand{\wg}[1]{}

\newcommand{\newtext}[1]{\textcolor{black}{#1}}

\begin{document}
\maketitle
\begin{abstract}
Unsupervised automatic readability assessment (ARA) methods have important practical and research applications (e.g., ensuring medical or educational materials are suitable for their target audiences). \newtext{In this paper, we propose a new zero-shot prompting methodology for ARA and present the first comprehensive evaluation of using large language models (LLMs) as an unsupervised ARA method by testing 10 diverse open-source LLMs (e.g., different sizes and developers) on 14 diverse datasets (e.g., different text lengths and languages). Our findings show that our proposed prompting methodology outperforms prior methods on 13 of the 14 datasets. Furthermore, we propose LAURAE, which combines LLM and readability formula scores to improve robustness by capturing both contextual and shallow (e.g., sentence length) features of readability. Our evaluation demonstrates that LAURAE robustly outperforms prior methods across languages, text lengths, and amounts of technical language.}
\end{abstract}







\section{Introduction}
\label{sec:intro}

Measuring the readability of text has many practical applications. For example, it is important when selecting or creating resources for educating school children~\citep{education_gpt} and second language learners~\citep{cambridge}. It is also critical in ensuring that patients who are diagnosed with a new disease or preparing to undergo treatment can understand, process, and act upon health-related information~\citep{americanheart,ophthalmology}. Healthcare institutions such as the American Medical Association and the National Institutes of Health provide recommendations for the readability level of patient materials to best improve patient outcomes~\citep{rooney21}. 

\newtext{Readability measurement has long been an important component of academic research as well. For example, prior work has used readability measures to ensure that resources are understandable for target audiences~\cite{jmir}, to validate text simplification or summarization methods~\cite{PICTON2025105743, wu2025rlpf}, and to examine document readability as a predictor of future outcomes (e.g., financial report readability and subsequent stock performance)~\cite{ZHANG2025108489}.}

 

\newtext{The time and costs required to accurately label the readability level of a large text corpus inspired the development of automatic readability assessment (ARA) tools. The earliest ARA tools were formulas that measured readability as a linear combination of shallow text features, such as average sentence length and syllables per word~\cite{fre}. The development of machine learning (ML), and specifically attention-based transformer models, led to the development of supervised ARA tools that combined text embeddings and linguistic features~\cite{cambridge,lee21}. While supervised methods substantially improved accuracy over traditional formulas, they require annotated corpora, technical expertise, and additional computational resources, which has limited their adoption. Most recently, researchers have proposed unsupervised language model-based ARA tools to balance the accuracy of deep learning methodologies with the usability of readability formulas~\cite{trott-riviere-2024-measuring,martinc21}.} 

However, we find that current research still heavily relies on readability formulas. \newtext{We  search the \textit{Scopus} database for peer-reviewed papers published in 2025 
with ``readability'' and at least one of the words ``formula'', ``NLP'', ``LLM'', ``ML'', ``AI'', or ``BERT'' in the titles, keywords, or abstract. We manually identify the papers that apply ARA tools (excluding proposals for new methods). We find two papers using supervised BERT models, two papers with unsupervised measures based on pretrained language model's (PLM) surprisal, and five papers with zero-shot prompted LLMs. In contrast, we find 337 papers using readability formulas. Readability formulas are applied across a diverse set of medical~\cite{Uysal2025}, linguistics~\cite{Bao2025}, business~\cite{Chabot2025}, AI~\cite{hossen2025unmasking}, and governance~\cite{Raman2025} research papers, and in many reputable conferences and journals, including the \textit{AAAI Conference on Artificial Intelligence}~\cite{wu2025rlpf}, \textit{Scientific Reports}~\cite{Uysal2025}, \textit{PLoS ONE}~\cite{Kacer2025}, \textit{Finance Research Letters}~\cite{Pathak2025}, and \textit{Journal of Medical Internet Research}~\cite{jmir}. }


\newtext{Although some delay may be expected, we believe that there are legitimate reasons why unsupervised language model-based ARA methods have not yet replaced readability formulas.~\citet{trott-riviere-2024-measuring} first showed that zero-shot prompting of GPT-4 models outperformed traditional readability formulas on one English dataset. While this limited evaluation shows the potential of the method, it may also inhibit adoption as researchers do not know if the performance generalizes to technical texts (e.g., medical resources), non-English languages, or when using free open source models.}
In this paper, we make \newtext{four} contributions. \newtext{First, we propose two methodological advances for using LLMs for unsupervised ARA: 1) we compute readability scores as an expected value over the output token probabilities, and 2) we prompt the model on the same scale used by manual annotators and include detailed definitions of each readability level when available. 
Experimental results show that prompting LLMs with our two proposed techniques leads to improved performance in comparison to~\citet{trott-riviere-2024-measuring} on 13 of 14 datasets, with particularly strong gains on non-English datasets.
}

Second, we propose a novel readability assessment method called \textbf{L}LM-based \textbf{A}utomatic \textbf{U}nsupervised \textbf{R}eadability \textbf{A}ssessment \textbf{E}nsemble (LAURAE), which combines zero-shot prompted LLM scores \newtext{and readability formula scores, based on weights derived from the LLM's stated confidence in its rating}. By considering both high-level contextual understanding from  LLMs and  shallow features (e.g., sentence length \newtext{and number of polysyllabic words) from readability formulas} to produce a  holistic readability assessment, \newtext{LAURAE outperforms readability formulas on all 14 datasets and standalone LLMs on 11 datasets.} 
%

Third, we present a more comprehensive evaluation of LLM-based methodologies for unsupervised ARA. Specifically, our evaluation is performed on 10 open-source LLMs with varying sizes, developers, and multilingual capabilities, and 14 datasets that vary in language, text length, and ground truth type.  \newtext{The evaluation shows that LAURAE is generalizable and an effective unsupervised ARA tool regardless of text length, language, and amount of technical content. Our finding that using the zero-shot methodology proposed in~\citet{trott-riviere-2024-measuring} for open-source LLMs only outperforms the best readability formula on 6 of 14 datasets, highlights the importance of our more thorough evaluation.}

Fourth, all of the code and datasets, including two Greek textbook datasets we curated, are publicly available\footnote{https://github.com/rag24/LAURAE}, providing resources to practitioners who need readability assessment tools and researchers for future work. 

In conclusion, our evaluation supports the adoption of \newtext{LAURAE for unsupervised ARA in practice or research. In particular, LAURAE is useful whenever high-quality readability assessment for a corpus of texts is desired and manual annotation is not plausible or too costly. Moreover, our findings show that combining contextual features from LLMs with shallow features from readability formulas enhances robustness in unsupervised ARA. Future research should explore whether combining zero-shot LLM ratings with shallow unsupervised NLP tools can further increase performance and robustness in other tasks such as sentiment analysis or toxicity detection.}
\section{Related Work}

In this section, we review automatic readability assessment (ARA) methods. Early work in this area developed traditional readability formulas that rely on shallow text characteristics such as average sentence or word length, the number of technical words, and the number of polysyllabic words. While many of these formulas were created decades ago~\citep{fre,ari}, new research continues in this area for languages that were previously unsupported~\citep{laurs-2024-towards,zhu24}. Traditional readability formulas are widely used since they are easy to implement and require no training data. However, as demonstrated in Section~\ref{sec:laurae_experiments}, their effectiveness is limited.

The next phase of ARA extracted linguistic features (e.g., number of noun phrases) from a text, as inputs to supervised machine learning (ML) models~\citep{cambridge,chatzipanagiotidis21,vajjala-meurers-12}. The availability of attention-based pretrained language models (PLMs) led to new methods that finetuned BERT-based models for ARA~\citep{martinc21}, including those which augmented BERT representations with extracted linguistic features~\citep{li22,imperial21,deutsch20,hou22,lee21}. Although these supervised approaches improved performance, they are often not applicable due to the costs of obtaining a large manually labeled training dataset.

Towards solving this issue,~\citet{lee22} proposed a neural pairwise ranking model, which can be applied to unseen datasets through transfer learning when the target and training datasets are very similar.~\citet{martinc21} proposed the ranked sentence readability score (RSRS), an unsupervised method that utilizes PLMs to quantify sentence-level readability by calculating the unexpectedness of each word. However, as shown in Section~\ref{sec:laurae_experiments}, RSRS does not consistently outperform readability formulas.

The most related work to ours is~\citet{trott-riviere-2024-measuring}, which proposes zero-shot prompting GPT-4 Turbo and GPT-4o for ARA. However, the evaluation is limited to a single prompting technique and English dataset. Another recent work zero-shot prompts closed source GPT models and the ChatGLM2-6B and Meta-Llama-3-8B models to assess readability on English and Chinese sentences. However, the lack of experimentation with prompts and larger open-source models results in poor performance~\citep{liu_automatic_2025}. 

In this work, we first address the limitations in prior works by proposing a new methodology for zero-shot prompting LLMs for unsupervised ARA, and evaluating this methodology when applied to 10 open-source LLMs and 14 diverse benchmark datasets. Furthermore, we propose a new method, LAURAE, that combines zero-shot prompting of LLMs with readability formula scores, and demonstrates robust and generalizable performance.

\section{Methodology}
\label{sec:method}
\newtext{We first propose a new method for obtaining unsupervised ARA scores from large language models (LLMs) in Section~\ref{sec:method_zeroshot}.} Then, in Section~\ref{sec:method_scores}, we introduce our proposed method, LAURAE, which ensembles the LLM readability scores from Section~\ref{sec:method_zeroshot} with \newtext{readability formulas scores, using weights based on the LLM’s verbalized confidence in its rating.}

\subsection{Prompting Zero-shot LLM}
\label{sec:method_zeroshot}

 \newtext{We propose two changes to the methodology for prompting LLMs in an unsupervised ARA task. First, we investigate whether or not including a definition of the readability scale in the prompt (when available) improves LLMs' zero-shot performance. Second, we calculate expected
  readability scores  as an expected value over the models’ output token probabilities, following prior research on using LLMs as raters~\cite{liu-etal-2023-g,lai2025beyond}. This is in contrast with the original proposal and current state-of-the-art~\cite{trott-riviere-2024-measuring}, which does not experiment with different prompts and considers only the number in the generated output text when producing a readability score.
}


\subsubsection{Readability Scale Definitions}

\newtext{
Human annotators either explicitly, or implicitly (e.g., school textbook datasets), rated the readability of the texts in each benchmark dataset. These ratings are used as the ground-truth readability scores, and for 7 of the 14 datasets, human raters relied on the Common European Framework of Reference for Languages (CEFR) to produce ratings. CEFR provides six levels of language proficiency, defined by the capabilities of learners at each level. For the datasets where manual raters used the CEFR scale and definitions, we prompt the LLM to generate scores on the same scale and include the definitions in the prompt. The six levels of CEFR proficiency (A1-C2) are converted to integers (1-6).}

\newtext{For the remaining datasets, we use a similar prompting approach to the existing literature~\cite{trott-riviere-2024-measuring}. We prompt the model for a readability score on an arbitrary scale (i.e., whole number value between 1 and 9) based on several key factors (e.g., grammar and clarity) as well as the models' own definition of readability. For comparison, we also test this prompting approach for the CEFR datasets. We display the actual prompts for each dataset in Appendix~\ref{sec:prompts}. }


\subsubsection{Expected Value of Output Tokens}
\label{subsubsec:expected_val_method}

\newtext{
  Instead of only considering the generated output token as in~\citet{trott-riviere-2024-measuring}, we calculate
  readability scores as an expected value over the models’ output token probabilities.}

\newtext{Formally, for the $i$-th generated token, the LLM produces a logit vector $z_i \in \mathbb{R}^{|V|}$, where $V$ is the LLM vocabulary. Applying the softmax function to $z_i$ yields probability distribution $p_i$, such that each element $p_{ij}$ corresponds to the probability of generating the $j$-th vocabulary token $V_j$, as the $i$-th token in the sequence. Let $n$ be the position in the generated text where the readability score is located. We create a new array $r$, with two rows and $|V|$ columns such that $r_{1k}$ equals the $k$-th highest probability in $p_n$, and $r_{2k}$ is the corresponding token index $j$, so that $p_{nj}=r_{1k}$.}

\newtext{In a zero-temperature setting, the model will generate the highest ranked-token, $V_{r_{21}}$. This is referred to as the vanilla score~\cite{lai2025beyond}. To calculate the expected value score, we first check whether $V_{r_{21}}$ is numeric and proceed to the next highest-ranked token until $V_{r_{2k}}$ is no longer numeric. The expected value score, $s_{LLM}$, is then computed as: }

\begin{equation}
    \label{eq:expected_val}
  s_{LLM}=   \sum_{m=1}^{k-1} r_{1m}(V_{r_{2m}})
\end{equation}

\subsection{LAURAE}
\label{sec:method_scores}

We propose \newtext{a new unsupervised automatic readability assessment (ARA) method that ensembles ratings from shallow unsupervised ARA techniques} and zero-short prompting of instruction-tuned LLMs. \newtext{In many cases, especially with longer or more technical texts that are heavily context-dependent, we expect zero-shot LLMs to outperform shallow unsupervised ARA methods. 
However, LLMs may struggle to rate the readability of very short texts that lack contextual information, or types of text that they are not well-suited for, such as children's stories~\cite{bhandari-brennan-2023-trustworthiness,valentini-etal-2023-automatic}. In these cases, unsupervised ARA methods such as readability formulas that focus on shallow features (e.g., sentence length or the number of polysyllabic words), may perform well. Thus, combining the scores of a shallow ARA method with the scores of a zero-shot LLM may improve overall performance by increasing robustness and generalizability.}  


\newtext{Since we are operating in an unsupervised setting, we can not use hyperparameter tuning to select the optimal weights for the two readability scores in the ensemble. Instead, we propose using the LLM's self-reported confidence to determine the weights. Prior work has shown that, compared to entropic uncertainty measures, LLMs produce more reliable and accurate confidence scores when prompted in natural language~\cite{tian-etal-2023-just}. Thus, we prompt each model to state its confidence, on a 1-9 scale, that the generated readability score will align with human raters' scores (full prompts in Appendix~\ref{sec:prompts}). We again apply the expected value scoring technique from Section~\ref{subsubsec:expected_val_method} to the LLM's confidence score, and divide the score by 10, to obtain our confidence weight $c$. The 1-9 scale ensures that each ARA method receives at least 0.1 weight in the final score. Given the LLM's readability score, $s_{LLM}$, and the readability formula's score, $s_{rf}$, we compute LAURAE's readability score as} 

\begin{equation}
    c(\frac{s_{LLM}-\mu_{LLM}}{\sigma_{LLM}}) + (1-c)(\frac{s_{rf}-\mu_{rf}}{\sigma_{rf}}),
    \label{eq:laurae}
\end{equation}

\newtext{where $\mu_{LLM}$, $\sigma_{LLM}$, $\mu_{rf}$, and $\sigma_{rf}$ are the dataset-level means and standard deviations of the respective readability scores.} 

\begin{table}[t!]
\resizebox{\linewidth}{!}{
\begin{tabular}{|>{\centering\arraybackslash}m{0.335\linewidth}|>{\centering\arraybackslash}m{0.2\linewidth}|>{\centering\arraybackslash}m{0.085\linewidth}|>
{\centering\arraybackslash}m{0.15\linewidth}|>{\centering\arraybackslash}m{0.23\linewidth}|}
\hline
\textbf{Dataset} & \textbf{Language} & \textbf{N} & \textbf{Avg. Length}  & \textbf{Ground Truth} \\
\hline
\multirow{5}{*}{ReadMe} 
& English & 296 & 22  & \multirow{7}{.1\textwidth}{\centering CEFR Rating} \\ \cline{2-4} 
& French & 185 & 25 &   \\ \cline{2-4} 
& Hindi & 163 & 23 &   \\ \cline{2-4} 
& Arabic & 206 & 24 &   \\ \cline{2-4} 
& Russian & 178 & 23 &   \\ \cline{1-4} 
MedReadMe & English & 1140 & 25  &  \\
\cline{1-4}
Cambridge & English & 300 & 579  &  \\
\hline
CLEAR & English & 1890 & 201  &  \multirow{4}{.11\textwidth}{\centering non-CEFR Rating} \\
\cline{1-4}
Greek Language & Greek & 393 & 161 &  \\ \cline{1-4} 
Greek History & Greek  & 804 & 209  &  \\ 
\cline{1-4}
OneStop & English & 567 & 782  &  \\
\hline
Asset & English & 485 & 21  & \multirow{3}{*}{\centering Comparison} \\
\cline{1-4}
\multirow{2}{*}{Vikidia} 
& English & 150 & 596  &  \\ \cline{2-4}
& French & 150 & 509  &  \\
\hline
\end{tabular}}
\caption{Datasets Used in Evaluation}
\label{tab:datasets}
\end{table}

\section{Experimental Setup}
\label{subsec:experiments_setup}


We start by introducing the experimental setup used to evaluate the zero-shot prompting ARA capabilities of LLMs (Section~\ref{sec:llm_eval}) as well as our proposed ensemble method, LAURAE (Section~\ref{sec:laurae_experiments}). Prior work on unsupervised ARA has yet to evaluate LLMs' ARA capabilities on diverse texts (e.g., differing text lengths and languages). Furthermore, the capabilities of leading open-source LLMs have not been tested, which may limit adoption. Our experimental setup aims to more comprehensively investigate whether LLMs can replace traditional readability formulas as the dominant unsupervised ARA method.

\textbf{Datasets} \newtext{We utilize the publicly available ReadMe~\cite{naous-etal-2024-readme}, MedReadMe~\cite{jiang-xu-2024-medreadme}, Cambridge~\cite{cambridge}, CommonLit Ease of Readability (CLEAR)~\cite{crossley23}, OneStop~\cite{onestop}, Asset~\cite{asset}, and Vikidia~\cite{vikidia} datasets. The 14 selected datasets and their key characteristics are shown in Table~\ref{tab:datasets}. Due to the limited availability of datasets with non-English texts longer than a sentence, we curated two additional datasets from publicly available Greek textbooks (see Appendix~\ref{sec:method_greek}).} The datasets vary across five key dimensions: 1) there are six different languages considered; 2) there are seven datasets with sentence-length texts (i.e., $\leq$25 words on average) and seven datasets with paragraph- or article-length texts;
3) there are seven datasets with readability scores and definitions based on the Common European Framework of Reference for Languages (CEFR) scale, 4) the MedReadMe dataset contains healthcare texts, which allows for evaluation of ARA on texts with technical language, and 5) there are eleven datasets with ground truth readability scores for each text and three datasets where the ground truth indicates which of two comparable texts is more readable. 


\yc{why only evaluate our method on MedReadMe, not all methods?}
\yc{together with the "Additionally,.." sentence, describe the domain of all datasets. }
\yc{I see the three paragraphs in Sec 3.2 on dataset definition fit better here, I suggest to shorten and move the content from sec 3.2 to the above}
\yc{I  suggest the discussion of dimensions in the order of 1) domain,  2) language,  3) lengths, 4) availability of readability definition, 5) availability of ground truth (with 2 variations), }

\yc{I am confused, is the 2 greek datasets included in the 14 datasets? if yes, move the above paragraph after the first sentence (ending "on 14 datasets", say sth like, in addition to public dataset available, we curated two datasets based on ...). if not, change 14 to 16 dataset, and report performance of all 16.}

\textbf{Evaluation Metrics.} For the 11 datasets that provide ground truth ratings, we report the Pearson correlation between each method's readability scores and the ground truth ratings, following the existing literature~\citep{naous-etal-2024-readme,jiang-xu-2024-medreadme,martinc21,trott-riviere-2024-measuring}. For the three datasets that provide ground truth comparisons, we report how accurately each method can identify the more readable of the two texts. \newtext{To test for statistically significant differences, we use Steiger's modification of Williams's test~\cite{steiger1980tests} to compare correlations and McNemar's test~\cite{McNemar_1947} to compare accuracies. }

\textbf{Open Source LLMs.}
We evaluate the ARA capabilities of a diverse set of open-source, instruction-tuned LLMs: \emph{Llama 3.1 8B/70B} and \emph{Llama 3.2 3B}~\cite{llama3modelcard}; the \emph{Aya Expanse} series~\cite{aya}; the \emph{Gemma 2} series~\cite{gemma}; \emph{Mixtral 8x7B}~\cite{mixtral} and \emph{Phi-4}~\cite{phi4}. Models vary from being monolingual (e.g., Phi-4 and Gemma 2 series) to specifically designed for multilingual capabilities (e.g., Aya Expanse series). Model size also varies from 2 billion (i.e., Gemma 2B) to 70 billion (i.e., Llama 70B) parameters. 

\subsection{Baseline Methods}
\label{sec:baselines}

\textbf{Readability Formulas.}
For English texts, we evaluate \emph{FKGL}~\cite{kincaid1975derivation} and \emph{ARI}~\cite{ari} due to their popularity and use in related works~\citep{naous-etal-2024-readme, jiang-xu-2024-medreadme}. 
We also evaluate \emph{OSMAN}~\cite{el-haj-rayson-2016-osman} for Arabic text, 
\emph{Lix}~\cite{anderson1983lix} for Hindi and Greek texts, and the adapted versions of \emph{Flesch Reading Ease (FRE)} for French and Russian texts \newtext{using the \textit{textstat}\footnote{https://pypi.org/project/textstat/} Python package.}



\textbf{RSRS.}~\citet{martinc21} proposed the Ranked Sentence Readability Score (RSRS) as a pretrained language model-based (PLM) unsupervised ARA method that uses a PLM's quantification of word unexpectedness and traditional shallow readability indicators such as sentence length (details in Appendix~\ref{sec:method_rsrs}). \newtext{We tested several variations (see Appendix~\ref{sec:xlmr}) and present results for RSRS implemented with a multilingual BERT model (\emph{mBERT})~\citep{devlin19}.}


\newtext{\textbf{LLM-v-ns.} A methodology for zero-shot prompting LLMs for ARA that uses vanilla readability scores (instead of expected value readability scores discussed in  Section~\ref{subsubsec:expected_val_method}) and only prompts LLMs for readability scores on an arbitrary scale. This is most similar to the method proposed in~\citet{trott-riviere-2024-measuring}.}


\begin{figure*}[t!]
\centering
\includegraphics[width=\textwidth]{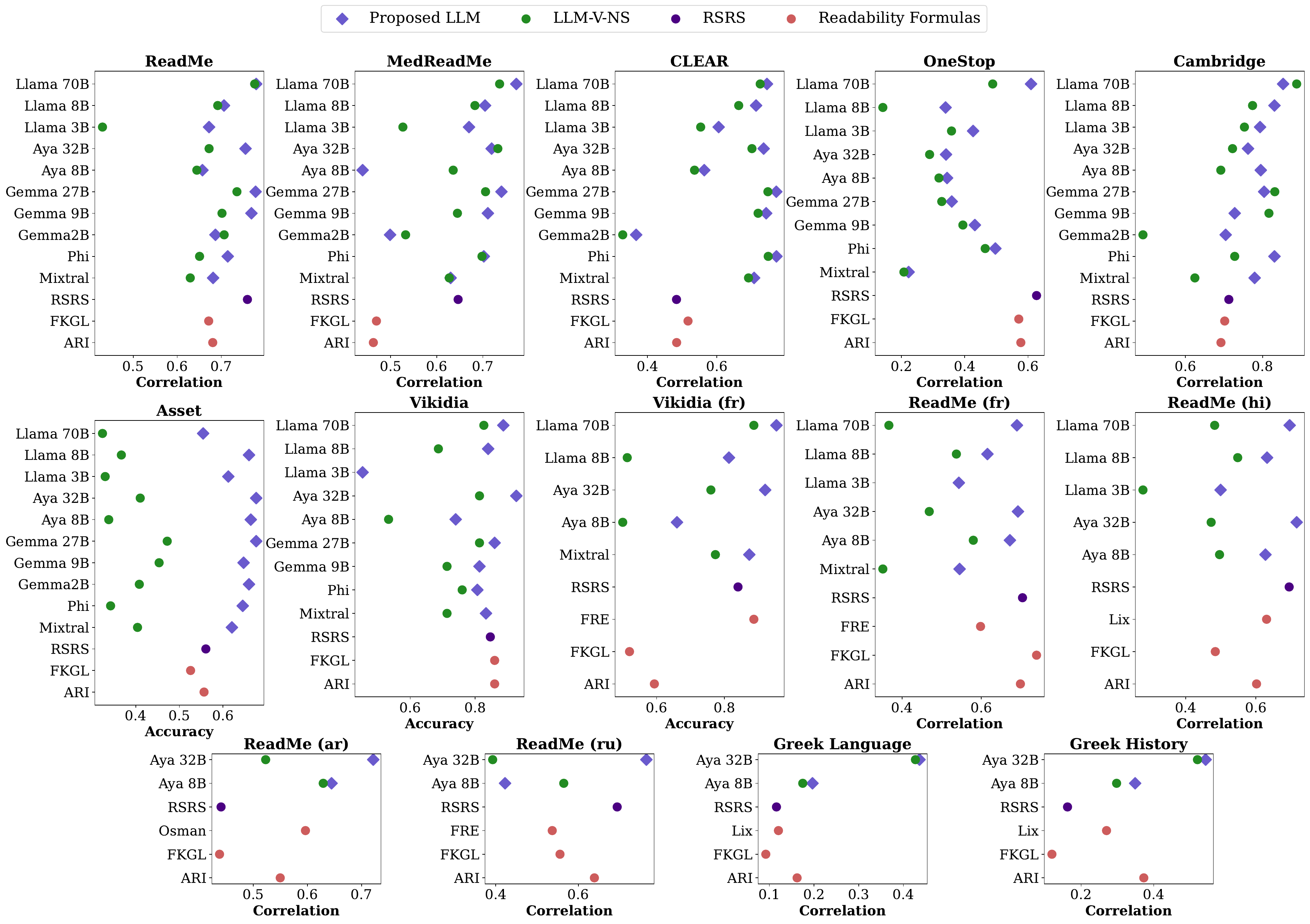}
\caption{\centering Proposed Zero-shot LLM Method versus Unsupervised ARA Baselines on Benchmark Datasets (Note: observations performing worse than 0.5 points behind the best method on each dataset are removed)}
\label{fig:llm_results}
\end{figure*} 

\section{Zero-shot Performance of Open Source LLMs}
\label{sec:llm_eval}

\newtext{In this section, we investigate the performance of three groups of methods: (1) the shallow unsupervised ARA methods (i.e., RSRS and readability formulas), (2) the zero-shot ARA performance of 10 popular open-source LLMs with the LLM-v-ns prompting methodology, and (3) the performance of the same 10 LLMs with our proposed zero-shot prompting methodology that includes the CEFR scale in prompts (for the seven relevant datasets) and computes readability scores as the expected value over the model's output token probabilities. The performance of these methods on 14 diverse readability datasets is shown in Figure~\ref{fig:llm_results}.}




\subsection{Comparison among existing LLMs-based methods}
\label{subsubsec:compare_llms}

\newtext{All included LLMs have English language capabilities, and thus, we can compare all of the models' performances on the English datasets (i.e., the first seven datasets in Figure~\ref{fig:llm_results}). The Llama 70B model, the largest LLM tested, is the top-ranked LLM four times, second-ranked once, and third-ranked once. Phi, Gemma 27B, and Aya 32B each perform as the top-ranked LLM on one English dataset. We conclude that the Llama 70B model has the best overall performance of the tested LLMs on English datasets.}

\newtext{The only dataset that Llama 70B  performs poorly on is Asset, which was uniquely created by human raters selecting the simpler text between two short sentences output by text simplification tools. Llama 70B rates the two sentences as equally readable 13.61\% of the time, which does not count as a correct prediction. No other LLM does this. Upon inspection, we found that many of the sentence pairs differed by just one or two words, which suggests that rating the two sentences as equally readable is reasonable.}

\newtext{For the seven non-English datasets, we only compare the performance of LLMs that were trained on texts in the target language. For the two French datasets and one Hindi dataset, we compare the Llama and Aya Expanse models. Surprisingly, Aya 32B outperforms Llama 70B on two of the three datasets, despite being smaller in size, highlighting the benefits of a focused approach to developing multilingual capabilities~\cite{aya}. The four datasets shown in the last row of Figure~\ref{fig:llm_results} have languages that are only supported by the Aya Expanse series, and Aya 32B always outperforms Aya 8B. In summary, of the models tested, Aya 32B has the best performance on non-English datasets.}

\newtext{We use Llama 70B for English datasets and Aya 32B for non-English datasets in the rest of the paper to replicate a truly unsupervised setting where the LLM can not be chosen by testing performance on a validation dataset.}

\subsection{Evaluation of Our Proposed LLM-based Method}
\label{subsubsec:compare_llms_baseline}
\newtext{Now we present the evaluation of our proposed improvements to ARA with zero-shot prompted LLMs. Across 14 datasets, our proposed method is the best performer on 11 datasets. The exceptions are the OneStop dataset where RSRS outperforms our proposed method by 0.018 points in correlation, the Cambridge dataset where the LLM-v-ns outperforms our method by 0.035 points in correlation, and the French ReadMe dataset where the FKGL and ARI formulas, as well as RSRS, outperform our method by 0.006-0.047 points in correlation. Even in a fully unsupervised setting (i.e., Llama 70B for English, and Aya 32B for non-English, texts), our method still performs the best on 10 datasets. }

\newtext{We are especially interested in comparing our proposed zero-shot prompting methodology to LLM-v-ns. Regardless of LLM, our method generally outperforms the baseline LLM-v-ns. Even on the Cambridge dataset, where the best performance is achieved by Llama 70B with the LLM-v-ns method, our proposed method performs better for 7 of the 10 evaluated LLMs. Focusing only on the recommended LLMs from Section~\ref{subsubsec:compare_llms}, our method outperforms LLM-v-ns on 13 datasets (with statistically significant differences for all but the English ReadMe and Greek Language datasets). }


\subsubsection{Ablation Studies}
\label{subsubsec:llms_ablation}
\newtext{
Recall that our proposed method has two improved components: 1) prompting the LLM for scores using the same scale as manual labelers, and 2) conducting expected value scoring. 
In Table~\ref{tab:llm_ablation}, we isolate the effect of these two components for the recommended LLM (see Section~\ref{subsubsec:compare_llms}) on each dataset. The ``Expected Value'' column shows the performance differences that resulted from expected value scoring, as opposed to vanilla. The use of expected value scores increases model performance on all 14 datasets, with 12 being significant increases. Improvements are particularly large for the datasets with ground truth comparisons instead of ratings (i.e., Vikidia and Asset). This is likely due to the expected value scoring technique reducing ties when comparable texts receive the same vanilla readability score. }

\newtext{The ``Scale Included'' column shows the isolated performance differences due to prompting the model to rate readability on the CEFR scale. This evaluation only applies to the seven datasets with ground truth values on the CEFR scale. Inclusion of the CEFR scale increases performance significantly for five of the seven datasets, and leads to particularly large increases for the non-English datasets. This suggests that LLMs may have less stored information about readability in non-English languages, and therefore benefit more from a well-defined readability scale.}
\begin{table}[t!]
\resizebox{\linewidth}{!}{
\begin{tabular}{|>{\centering\arraybackslash}m{0.32\linewidth}|>{\centering\arraybackslash}m{0.36\linewidth}|>{\centering\arraybackslash}m{0.32\linewidth}|}
\hline
\textbf{Dataset} & 
\textbf{Expected Value} & 
\textbf{Scale Included} \\
\hline
Greek Lang.  & 0.009 & -  \\
\hline
Greek History & 0.022\textsuperscript{**} & -  \\
\hline
Vikidia (fr) &  0.160\textsuperscript{***} & -  \\
\hline
Vikidia &  0.060\textsuperscript{**} & - \\
\hline
Asset &  0.231\textsuperscript{***} & -  \\
\hline
CLEAR &  0.020\textsuperscript{***} & -  \\
\hline
OneStop &  0.121\textsuperscript{***} & - \\
\hline
MedReadMe &  0.014 \textsuperscript{***} &  0.026\textsuperscript{***}  \\
\hline
Cambridge &  0.032\textsuperscript{***} & - 0.059\textsuperscript{***} \\
\hline
ReadMe &  0.018\textsuperscript{**} & - 0.026\textsuperscript{*}  \\
\hline
ReadMe (fr) & 0.016 &  0.214\textsuperscript{***}  \\
\hline
ReadMe (hi) &  0.058\textsuperscript{***} & 0.204\textsuperscript{***}  \\
\hline
ReadMe (ar) & 0.029\textsuperscript{**} & 0.177\textsuperscript{***}  \\
\hline
ReadMe (ru) &  0.030\textsuperscript{**} &  0.339\textsuperscript{***}  \\
\hline
\hline
\textbf{Average} & \textbf{+ 0.059} & \textbf{+ 0.125}\\
\hline
\multicolumn{3}{r}{\textsuperscript{***}$p<0.001$,\textsuperscript{**}$p<0.01$, 
\textsuperscript{*}$p<0.05$ }
\end{tabular}
}
\caption{\centering Performance Differences when Including CEFR Scale in Prompt and Using Expected Value Scoring by Dataset}
\label{tab:llm_ablation}
\end{table}

\section{LAURAE}
\label{sec:laurae_experiments}

\newtext{In this section we evaluate the performance of our proposed ensemble method LAURAE. Section~\ref{sec:lauraeE} shows that  by combining LLM readability scores with readability formula scores, LAURAE improves the performance and generalizability of the LLM-based approach presented in Section~\ref{sec:llm_eval}. We perform two additional analyses to verify LAURAE's effectiveness of using the LLM's verbal confidence scores as ensemble weights  in Sections~\ref{subsec:calibration} and~\ref{subsec:laurae_ablation}. Finally, we investigate whether LAURAE can benefit from dataset-level ensemble weights in Section~\ref{subsubsec:dataset_weights}. }


\subsection{LAURAE Evaluation}
\label{sec:lauraeE}
\newtext{As shown in Table~\ref{tab:laurae_results}, LAURAE,  using readability formula scores, outperformed each of the three baselines. Although slightly worse in terms of overall performance, LAURAE still outperforms the baseline methods when RSRS replaces readability formulas as the shallow feature readability score (see Appendix~\ref{sec:laurae_rsrs}). Consistent with Section~\ref{subsubsec:compare_llms}, we use Llama 70B for English datasets and Aya 32B for non-English datasets. For simplicity, we select the best readability formula on each dataset.}


\newtext{The main takeaway from Table~\ref{tab:laurae_results} is that LAURAE outperforms all baselines on 13 of 14 datasets. Only the LLM-v-ns baseline method outperforms LAURAE on the Cambridge dataset. This may be attributed to its better performance than our proposed zero-shot prompting methodology on this dataset (see Section~\ref{subsubsec:compare_llms_baseline}). Notably, LAURAE reduces the performance gap between our proposed prompting methodology and LLM-v-ns on that dataset (see Figure~\ref{fig:laurae_ablation}). LAURAE's improved performance is significantly different from the performance of RSRS, LLM-v-ns, and readability formulas on 10, 11, and 12 of the datasets, respectively.
}

\yc{why "significantly different", not "significantly better"?}

\newtext{A second takeaway is that across the 14 datasets, all three baselines have a similar average performance. 
This is a surprising finding because prior work had shown that the LLM-v-ns methodology applied to GPT-4 outperformed traditional readability formulas~\cite{trott-riviere-2024-measuring}. However, they only studied the CLEAR dataset, for which our findings are consistent (Table~\ref{tab:laurae_results}). This new insight shows the importance of comprehensive evaluation, as we did with 14 diverse datasets.
}

\begin{table}[t!]
\resizebox{\linewidth}{!}{
\begin{tabular}{|>{\centering \arraybackslash}m{0.26\linewidth}|>{\arraybackslash}m{0.19\linewidth}|>{\arraybackslash}m{0.225\linewidth}|>
{\arraybackslash}m{0.18\linewidth}|>
{\arraybackslash}m{0.145\linewidth}|}
\hline
\textbf{Dataset} & 
\textbf{LAURAE} & 
\textbf{LLM-v-ns} & 
\textbf{Formula} & 
\textbf{RSRS}\\
\hline
Greek Lang.  & \textbf{0.43} & 0.427 & 0.162\textsuperscript{***} & 0.116\textsuperscript{***}  \\
\hline
Greek Hist. & \textbf{0.572} & 0.52\textsuperscript{***} & 0.373\textsuperscript{***} & 0.163\textsuperscript{***}  \\
\hline
Vikidia (fr) & \textbf{0.953} & 0.76\textsuperscript{***} & 0.887\textsuperscript{*} & 0.84\textsuperscript{***}  \\
\hline
Vikidia & \textbf{0.9} &  0.827\textsuperscript{***} &  0.86&  0.847\textsuperscript{*} \\
\hline
Asset &  \textbf{0.629} & 0.324\textsuperscript{***} & 0.557\textsuperscript{***} & 0.561\textsuperscript{**} \\
\hline
CLEAR & \textbf{0.735} & 0.725\textsuperscript{*} & 0.517\textsuperscript{***} & 0.484\textsuperscript{***} \\
\hline
OneStop & \textbf{0.654} & 0.488\textsuperscript{***} & 0.577\textsuperscript{**} & 0.627 \\
\hline
MedReadMe & \textbf{0.77} & 0.736\textsuperscript{***} &  0.469\textsuperscript{***}  & 0.646\textsuperscript{***} \\
\hline
Cambridge & 0.86 & \textbf{0.888}\textsuperscript{*} & 0.702\textsuperscript{***} & 0.713\textsuperscript{***} \\
\hline
ReadMe & \textbf{0.798} & 0.776 & 0.68\textsuperscript{***}& 0.759 \\
\hline
ReadMe (fr) & \textbf{0.75} &  0.469\textsuperscript{***}&  0.739 &  0.704  \\
\hline
ReadMe (hi) & \textbf{0.754} & 0.473\textsuperscript{***} & 0.631\textsuperscript{**} & 0.695  \\
\hline
ReadMe (ar) & \textbf{0.757} & 0.523\textsuperscript{***} & 0.596\textsuperscript{***} & 0.441\textsuperscript{***}  \\
\hline
ReadMe (ru) & \textbf{0.803} & 0.393\textsuperscript{***} &  0.639\textsuperscript{***}& 0.694\textsuperscript{***}  \\
\hline
\hline
\textit{Average} & \textbf{\textit{0.74}} & \textit{0.595}& \textit{0.599} & \textit{0.592} \\
\hline
\multicolumn{5}{r}{\textsuperscript{***}$p<0.001$,\textsuperscript{**}$p<0.01$, 
\textsuperscript{*}$p<0.05$ }
\end{tabular}
}
\caption{\centering Performance Evaluation of proposed LAURAE (Note: best result bolded and significance testing indicates difference from LAURAE result)}
\label{tab:laurae_results}
\end{table}



\subsection{Verbal Confidence Score Calibration}
\label{subsec:calibration}
Now we study the effectiveness of using the LLM’s verbal confidence
scores as ensemble weights in LAURAE.
Prior research has shown that LLMs are often overconfident in their predictions~\cite{tian-etal-2023-just,zhou-etal-2023-navigating}. To verify that verbal confidence scores are a useful proxy for LLM accuracy, we show that the correlation between LLM and ground-truth readability scores is stronger among ratings associated with higher LLM confidence scores in each dataset. For each of the 11 datasets with ground truth readability ratings (see Table~\ref{tab:datasets}), we compute the verbal confidence score associated with the 25th and 75th percentiles. We then separate out the texts corresponding to verbal confidence scores in the top and bottom quartiles of verbal confidence scores respectively. In all but one dataset (Greek History), the highest-confidence quartile showed substantially stronger correlations with the ground truth than the lowest-confidence quartile. On average, the difference between the top and bottom quartiles was 16.7 correlation points. The Greek History dataset had low performance in both quartiles (correlations of 0.1096 and 0.4108, respectively), indicating that the issue is with the overall model performance on this dataset, and not the verbal confidence scores. 

We only evaluate the usefulness of verbal confidence scores in the datasets with a ground truth rating because for those datasets with ground truth comparisons (i.e., Text 1 is simpler than Text 2), our method rates each text independently. In other words, the confidence scores only indicate how confident the LLM is in rating the readability of a single text, not in the comparison between two texts.

\subsection{LAURAE Ablation Studies}
\label{subsec:laurae_ablation}

\begin{figure}[t!]
\centering
\includegraphics[width=\linewidth]{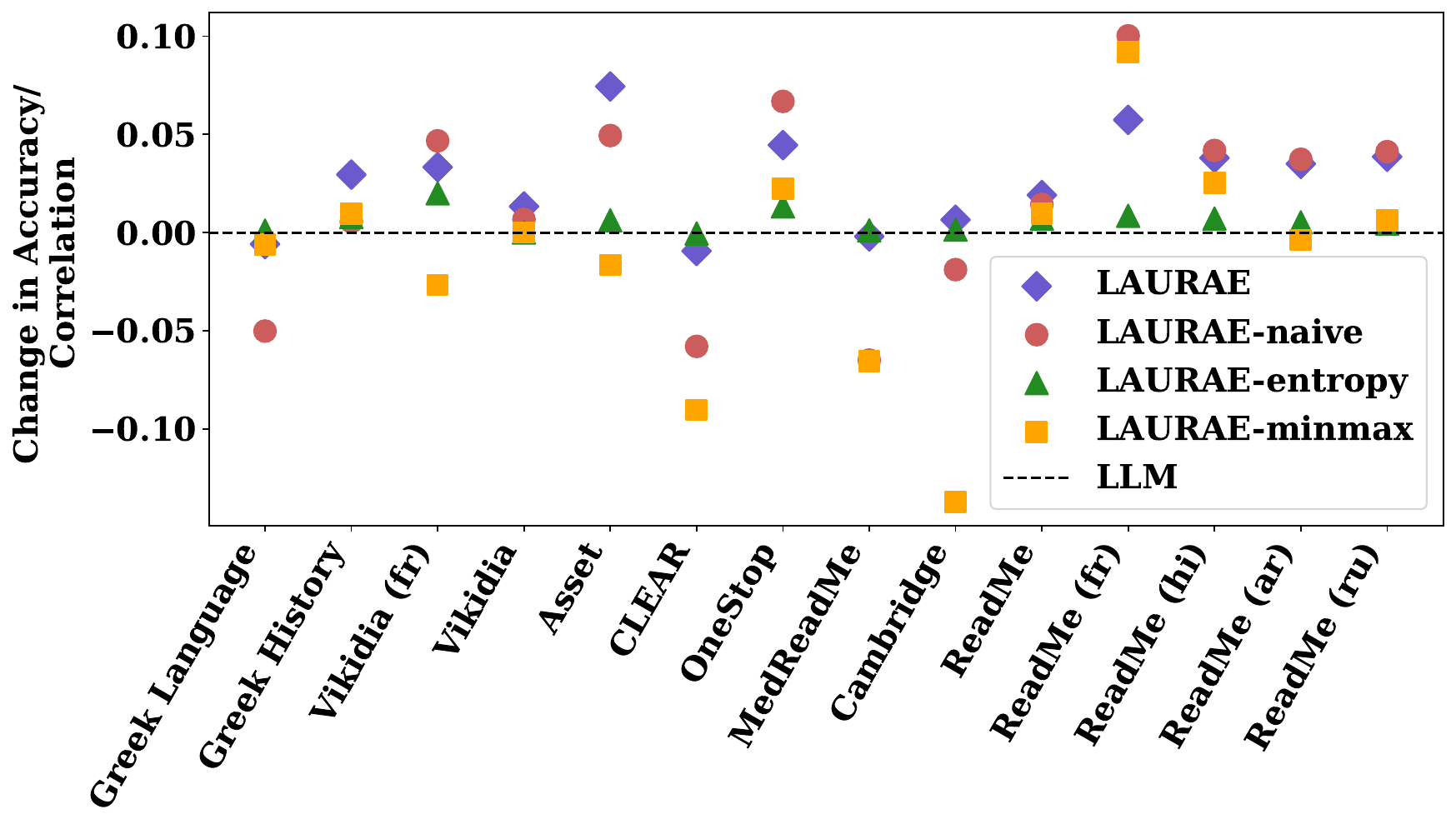}
\caption{LAURAE Ablation Study Results}
\label{fig:laurae_ablation}
\end{figure}

We compare the performance between LAURAE and the proposed standalone zero-shot LLM method in Figure~\ref{fig:laurae_ablation}. We also compare with three other ensemble variations to show the effectiveness of verbal confidence scores as weights: 
\begin{itemize}
    \item[1)] LAURAE-naive, a method that combines the standardized LLM and readability formula scores with equal weights;
    \item[2)] LAURAE-entropy, a method that derives an ensemble weight based on the Shannon entropy of the LLM's output distribution for the readability score token\footnote{Specifically, the LLM and readability formula weights in the ensemble are $1-H$, and $H$, respectively, where $H$ is entropy.}; and
    \item[3)] LAURAE-minmax, a method that applies min-max normalization at the dataset-level to ensure confidence scores are distributed across the entire interval $[0,1]$  based on relative confidence. 
    
\end{itemize}

Our findings show that combining the contextual LLM and shallow readability formula scores brings further improvements in comparison with our proposed zero-shot prompting methodology. Furthermore, we show that, on average, verbal confidence scores are the best technique for determining ensemble weights.

In Figure~\ref{fig:laurae_ablation}, the standalone LLM is considered as the baseline performance, and we plot differences in performance for LAURAE and its variants. Only the LAURAE-minmax variant performs worse than a standalone LLM, by -0.013 points on average.  LAURAE-entropy,  LAURAE-naive, and LAURAE all outperform a standalone LLM, by 0.006, 0.015, 0.027 points on average, respectively.


In the datasets where a standalone LLM is the best performing method (i.e. Greek Language and CLEAR datasets), LAURAE and LAURAE-entropy display robustness by keeping the differences small (i.e., less than 0.01 points in terms of accuracy/correlation). In contrast, LAURAE-naive performs at least 0.05 points worse than a standalone LLM on three datasets. 

In summary, empirical evaluation shows that the proposed LAURAE is the best performing variant due to its higher average performance, and demonstrated robustness. 

\subsection{Dataset-Level Weights}
\label{subsubsec:dataset_weights}

We test whether a variation on LAURAE that aggregates verbal confidence scores to the dataset-level to reduce noise can improve effectiveness.

It is clear from Table~\ref{tab:laurae_results} that LLMs are better suited to evaluate readability on some datasets (e.g., Cambridge and ReadMe) than others (e.g., Asset and Greek textbook datasets). One possible way to determine how useful LLM readability scores are for a dataset is to utilize the average verbal confidence score from all texts in the dataset (distribution of verbal confidence scores within each dataset is reported in Appendix~\ref{app:laurae_agg}). Given potential noises in individual text-level confidence scores, we test the effectiveness of a new LAURAE variant that replaces the individual text confidence score, $c$, in Eq.~\ref{eq:laurae} with the average confidence score from all texts in the same dataset. We call this variant LAURAE-agg. On average, LAURAE-agg improves performance by 0.0007 points in terms of accuracy/correlation across the 14 datasets, as reported in Appendix~\ref{app:laurae_agg}.

\section{Conclusion}
\newtext{In comparison to prior work, we more thoroughly investigate whether zero-shot prompting LLMs is an effective method for unsupervised ARA, by using 14 diverse benchmark datasets. 
This thorough evaluation yields outcomes that differ from those reported in narrow-scope studies in the literature~\citet{trott-riviere-2024-measuring}. It also reveals the need for language models specifically designed to have multilingual capabilities. }

\newtext{We also made methodological contributions. We propose two changes to the prompting methodology: 1) prompt the LLM for readability scores on the same scale used by manual labelers, 2) and compute readability as an expected value over the output token probabilities. We further propose LAURAE, a novel method that combines readability scores from LLMs and traditional formulas using weights determined by the LLM's confidence. This combination harnesses the strengths of standalone LLMs (e.g., ability to rate readability with respect to a specific definition), while also avoiding some of their pitfalls (e.g., low performance on unseen writing styles) for increased robustness.}


\newtext{LAURAE consistently outperforms existing unsupervised ARA methods by substantial margins. Besides supporting widespread adoption of LAURAE, these findings suggest that future research should investigate if combining zero-shot LLMs with shallow feature methods can benefit other unsupervised NLP tasks.
}


\section*{Acknowledgments}
Research reported in this publication was supported  in part by 
the National Center For Advancing Translational Sciences of the National Institutes of Health under Award Number UM1TR004789, the NSF under Grant No. CNS2237328 as well as by the Martin Tuchman'62 Chair Endowment and the Leir Foundation.  The content is solely the responsibility of the authors and does not necessarily represent the official views of the funders.
We thank the  Computer Technology Institute and Press ``Diophantus'' for providing open access to Greek textbooks. 

\section{Limitations}

There are two additional requirements for using our proposed method compared to using the traditional readability formulas: 1) Python literacy, and 2) access to computational resources.


First,  while minimal coding is required for readability assessment using traditional formulas, as Python packages like Textstat generate scores with a single line of code, Python literacy is required to use our proposed method. This may seem like a barrier for some users (e.g. healthcare professionals, and educators). We will try to alleviate this issue by providing detailed instructions along with all the code to aid users.
We hope users will ultimately weigh the performance gains of our method against its requirements when selecting an unsupervised readability assessment method.
\wg{The above paragraph can be shortened, the last sentence can be removed}


Second, computational resources are required for running LLMs. 
In this research, for any LLM with 14 billion or fewer parameters, we use one Nvidia A100 GPU for model inference. We use two Nvidia A100 GPUs for inference with Aya Expanse 32B, and three Nvidia A100 GPUs for inference with the Mixtral and Llama 70B models. All variations of RSRS are performed with a single Nvidia A100 GPU. One possibility that can be explored in future research is whether performance degrades significantly with quantized versions of the LLM (e.g. 8-bit floating point precision instead of 16-bit like we use). If performance is comparable, this would reduce the demand for computational resources. 

One limitation of LAURAE in comparison to readability formulas is that LAURAE is not interpretable. The performance gap between the two methods is substantial enough that this is not a reason to use readability formulas; however, it should encourage future research to investigate the generated natural language explanations from LLMs prompted to perform readability assessment. As shown in Appendix~\ref{sec:prompts}, we do prompt the model to explain its output after providing readability and confidence scores. In this current work, we did not evaluate the plausibility or faithfulness of the explanations, because the focus was on improving the performance of unsupervised ARA methods. However, future research should explore whether these explanations are useful. If the explanations are useful (or can be made useful through changes to the prompting methodology), it would allow for an interpretable version of LAURAE because both components would be interpretable.

Similarly, the evaluation is based on correlation with ground truth readability scores rather than accuracy of absolute readability scores (e.g., 5th grade-level or A2 CEFR level). Though we do not utilize them in this way, LLMs can simply choose a grade or CEFR level readability score. Future research should LLMs absolute readability scoring accuracy and whether ensemble approaches offer improvements in this area.

An additional minor limitation of our paper is that the evaluation is limited to the datasets we could obtain from past papers or curate ourselves. Thus, we are only able to test our paper against one medical dataset (i.e., MedReadMe). Similarly, we were unable to find other open-access textbook repositories to test whether the poor performance of our method on the Greek textbooks was a byproduct of the limited training resources available for Greek, or if it is a limitation of our method (e.g., RSRS struggles to perform well on textbooks targeted to children).

\section{Ethical Considerations}

We do not condone the use of our method for evaluating individuals' writing. Although the method outperforms traditional readability formulas, it does not have perfect accuracy and could even display biases against certain writing styles. Our method is intended for use in applications where a collection of texts need to be automatically rated for readability. For example, it can be used to compare medical resources for patients and determine which texts are more readable. It may also be used to test the effectiveness of text simplification methods. 

Second, LLMs are associated with high energy costs. Typically papers report the energy costs and environmental harm of training LLMs. 
Our proposed method does not require additional training, but uses model inferences \citep{energycosts}, which still have some energy costs. We encourage users to take these costs into consideration when choosing an unsupervised ARA method.

\bibliography{custom, anthology}

\appendix

\section{Prompts}
\label{sec:prompts}

We report the exact prompts used for each of the datasets in our study. \newtext{There are two variations of the prompts. For the seven datasets (Cambridge, MedReadMe and five ReadMe datasets) where ground truth ratings are based on the CEFR scale, we prompt the model to output a score on the CEFR scale and provide definitions of each level. For the remaining seven datasets, we prompt the model to produce readability scores on an arbitrary 1-9 scale, considering a few specific factors of readability (e.g., sentence structure and grammar), and the model's own definitions of readability. For comparison, we do prompt the models to provide a readability score on the arbitrary scale for the seven CEFR datasets too.}

In the early stages of this project, we revised the prompts several times to ensure that the models consistently output numbers on the requested scale (e.g., 1-9). Ultimately, we settled on requesting a specific format for the model output to ensure that we could easily extract readability scores from each model output. We also evaluated the open-source Falcon-7B-Instruct model in our initial experiments; however, it was excluded from the final experiments because we were unable to find a prompt that encouraged the model to reliably return numeric readability scores.

\subsection{CEFR prompts}

All of the five ReadMe datasets and the MedReadMe dataset have the same prompt because the ground-truth ratings are on a 6-point scale based on CEFR levels. \newtext{The CEFR level definitions are from ~\citet{naous-etal-2024-readme}. For all LLMs, the prompt is: }

\newtext{\textit{Rate the readability of the text between 1 (very easy) and 6 (very challenging) using the following scale: 1 = Can understand very short, simple texts a single phrase at a time, picking up familiar names, words and basic phrases and rereading as required. 2 = Can understand short, simple texts on familiar matters of a concrete type 3 = Can read straightforward factual texts on subjects related to his/her field and interest with a satisfactory level of comprehension. 4 = Can read with a large degree of independence, adapting style and speed of reading to different texts and purpose 5 = Can understand in detail lengthy, complex texts, whether or not they relate to his/her own area of specialty, provided he/she can reread difficult sections. 6 = Can understand and interpret critically virtually all forms of the written language including abstract, structurally complex, or highly colloquial literary and non-literary writings. You may use both the provided scale and your own understanding of readability to determine the most appropriate score. Where on the scale of readability does this text rate: [INSERT TEXT]. Additionally, state how confident you are that your rating will align with human raters, with a whole number value between 1 and 9. Answer with this format: Answer: [SCORE] Confidence: [Confidence Score] Explanation: [EXPLANATION]}}

\subsubsection{Cambridge}

The Cambridge dataset is also based on the CEFR scale, but there are no texts at the easiest level (i.e., A1), so we adjust the prompt slightly: 

\newtext{\textit{Rate the readability of the text between 1 (easy) and 5 (very challenging) using the following scale: 1 = Can understand short, simple texts on familiar matters of a concrete type 2 = Can read straightforward factual texts on subjects related to his/her field and interest with a satisfactory level of comprehension. 3 = Can read with a large degree of independence, adapting style and speed of reading to different texts and purpose 4 = Can understand in detail lengthy, complex texts, whether or not they relate to his/her own area of specialty, provided he/she can reread difficult sections. 5 = Can understand and interpret critically virtually all forms of the written language including abstract, structurally complex, or highly colloquial literary and non-literary writings. You may use both the provided scale and your own understanding of readability to determine the most appropriate score. Where on the scale of readability does this text rate: [INSERT TEXT]. Additionally, state how confident you are that your rating will align with human raters, with a whole number value between 1 and 9. Answer with this format: Answer: [SCORE] Confidence: [Confidence Score] Explanation: [EXPLANATION].}}

\subsection{Arbitrary Readability Scale Prompts}

\newtext{All datasets are prompted to generate a readability score based on an arbitrary scale where 1 indicates the text is very easy to understand and 9 indicates the text is very difficult to understand. Several of the benchmark datasets have texts that were not explicitly rated by manual annotators, but were implicitly rated based on comparisons to another text (e.g., Asset or Vikidia datasets) or the text being written to a specific audience (e.g., Greek textbook datasets). For these datasets, we append an additional short description about where the texts are sourced from. This description is added to the beginning of the prompt. We include these descriptions after the base prompt.}

\subsubsection{Base Prompt}

\newtext{\textit{Rate the readability of the text with a whole number value between 1 (very easy to understand) and 9 (very difficult to understand). Consider factors such as sentence structure, vocabulary or grammar complexity, and overall clarity, as well as your own understanding of readability.  Where on the scale of readability does this text rate: '{s}'. Additionally, state how confident you are that your rating will align with human raters, with a whole number value between 1 and 9. Answer with this format: Answer: [SCORE] Confidence: [Confidence Score] Explanation: [EXPLANATION]}} 

\subsubsection{Greek Language Textbooks}

The Greek Language textbooks dataset is based on textbooks that are designed for Greek schoolchildren between second and sixth grade, so we adjust the prompt to include this context: 

\newtext{\textit{Rate the readability of excerpts from Greek language textbooks targeted for students between second and sixth grades.}}

\subsubsection{Greek History Textbooks}

The Greek History textbooks dataset is based on textbooks that are suitable for Greek schoolchildren between fourth and twelfth grade, so we adjust the prompt to include this context:

\newtext{\textit{Rate the readability of excerpts from Greek History textbooks targeted for students between fourth and twelfth grades.}}

\subsubsection{Vikidia Datasets}

\newtext{Both Vikidia datasets are sourced from Wikipedia articles that were manually rewritten to be suitable for children audiences. do not have any readability scale. Thus, we adjust the prompt to include this context:}

\newtext{\textit{These Wikipedia articles are either intended for adult audiences or manually rewritten for children audiences.}}

\subsubsection{Asset Dataset}

\newtext{The Asset dataset was created by a manual evaluation of text simplification methodologies applied to short sentences. Thus, we adjust the prompt to include this context:}

\newtext{\textit{All of these sentences were rewritten to compare different text simplification methodologies.}}

\section{Greek Textbook Datasets}
\label{sec:method_greek}
We aim to evaluate the proposed method on datasets that vary in text length, text content, and language. However, one type of dataset we could not obtain\footnote{Although these datasets exist in prior research, we could not find a publicly available one that suited our needs and did not hear back from the authors we contacted.} was a non-English dataset with texts that were longer than a single sentence and a ground truth rating instead of a ground-truth comparison between two similar texts. 

One paper~\citep{greektexts} created a readability dataset from the open-access repository of Greek textbooks available at \url{http://ebooks.edu.gr/ebooks/}. Although we did not find this exact dataset, we were able to create a similar dataset from the repository. 

We first collected three language textbook PDFs for each grade of 2nd, 4th, and 6th graders  and the history textbook PDFs for 4th, 6th, 10th  and 12th grade. 
To extract a meaningfully long passage, we only considered passages which were at least ten lines long. 
\yc{ten lines long have nothing to do with meaningful. you need to motivation as sth like "Since long passages are more likely to present readability challenges"}
We manually removed any passages which were copyright information, author information, or lists of reading materials. We were left with 393 passages from the language textbooks and 804 passages from the history textbooks. The ground truth readability score of each passage is the grade level of the textbook the excerpt is from.

\section{RSRS}
\label{sec:method_rsrs}


Proposed by~\citet{martinc21}, the Ranked Sentence Readability Score (RSRS) is calculated at the sentence level, as defined below. Note that the RSRS of a document is the average RSRS of all its sentences. 
\begin{equation}
\text{RSRS} = \frac{\sum^s_{i=1}\sqrt{i} \times WNLL(i)}{s}
\end{equation}
where $s$ is the sentence length (i.e. the number of tokens in the sentence), and $i$ is the $i^{th}$ ranked token in terms of the smallest word-negative log likelihood (WNLL). 

To calculate the WNLL for a  token, we first mask the target token and pass the masked sentence to a pre-trained language model (e.g., BERT).
The language model makes a prediction for the masked token. Specifically, the output is a vector, $y_p$, whose length equals the vocabulary size. The $j^{th}$ value of $y_p$ represents the model's predicted probability of the masked token being the $j^{th}$ token in the vocabulary. 
The WNLL of a target token is then computed as:
\begin{equation}
\text{WNLL} = -(y_t\log{y_p}+(1-y_t)\log{(1-y_p)})
\end{equation}
where $y_t$ is a vector whose length equals the vocabulary size, and corresponds to the ground truth value for the target token (i.e., it has a 1 in the index position for the  ground truth token and 0s for all other positions). If the model expects the ground truth value with high probability, the WNLL will be small, otherwise (i.e. the token is unexpected), the WNLL will be large. 

\section{Selecting PLM for RSRS}
\label{sec:xlmr}

\begin{table}[t!]
\resizebox{\linewidth}{!}{
\begin{tabular}{|>{\centering \arraybackslash}m{0.26\linewidth}|>{\arraybackslash}m{0.21\linewidth}|>{\arraybackslash}m{0.21\linewidth}|>
{\arraybackslash}m{0.31\linewidth}|}
\hline
\textbf{Dataset} & 
\textbf{mBERT} & 
\textbf{XLM-R} & 
\textbf{Monolingual} \\
\hline
Greek Lang.  & 0.116 & 0.112 & \textbf{0.159}  \\
\hline
Greek Hist. & 0.163 & 0.138 & \textbf{0.191}  \\
\hline
Vikidia (fr) & 0.84 & 0.833 & \textbf{0.853}   \\
\hline
Vikidia & 0.847 & \textbf{0.867} & 0.847  \\
\hline
Asset & 0.561 & \textbf{0.563} & 0.532 \\
\hline
CLEAR & 0.484 & 0.428 & \textbf{0.54} \\
\hline
OneStop & \textbf{0.627} & 0.618 & 0.597 \\
\hline
MedReadMe & 0.646 & 0.567 &\textbf{0.68} \\
\hline
Cambridge & \textbf{0.713} & 0.656 & 0.65 \\
\hline
ReadMe & 0.759 & 0.746 & \textbf{0.782}  \\
\hline
ReadMe (fr) & 0.704 & \textbf{0.707} & 0.688   \\
\hline
ReadMe (hi) & \textbf{0.695} & 0.619 & 0.605 \\
\hline
ReadMe (ar) & 0.441 & \textbf{0.6} & 0.466   \\
\hline
ReadMe (ru) & \textbf{0.694} & 0.621 & 0.522\\
\hline
\hline
\textit{Average} & \textbf{\textit{0.592}} & \textit{0.577}& \textit{0.579}  \\
\hline
\end{tabular}
}
\caption{\centering Performance Evaluation of Pretrained Language Models for RSRS. Note: Best performance bolded.}
\label{tab:rsrs_ablation_results}
\end{table}

\newtext{For the RSRS method~\citet{martinc21}, we follow test three variants of RSRS\footnote{Our implementation of RSRS is based on code that was made available at: https://github.com/kinimod23/GRANT.} We implement RSRS with two multilingual BERT-based models: \emph{mBERT}~\citep{devlin19} and \emph{XLM-R}~\citep{xlmr}. We also choose one monolingual BERT-based model for each dataset depending on the language: \emph{BERT}~\citep{devlin19} for English, \emph{IndicBERTv2}~\cite{indicbert} for Hindi, \emph{ARBERTv2}~\cite{arbert} for Arabic, \emph{ruBert}~\cite{rubert} for Russian, \emph{CamemBERT(a)v2}~\cite{camembert} for French, and \emph{GreekBERT}~\cite{greekbert} for Greek. }

\begin{figure}[t!]
\includegraphics[width=\linewidth]{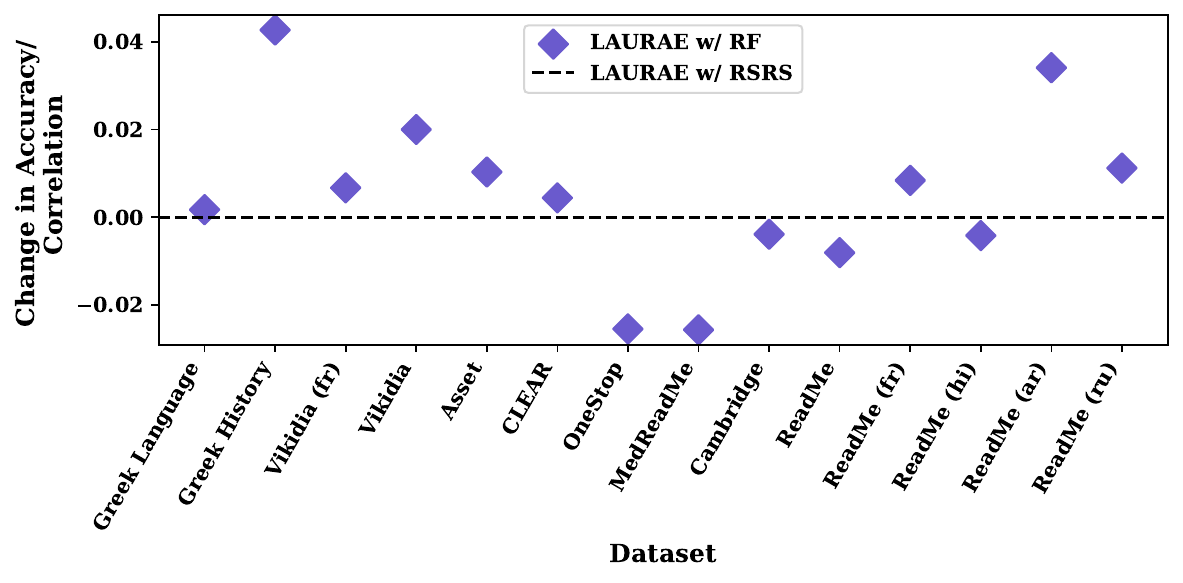}
\caption{\centering LAURAE Ablation Study for Shallow Unsupervised ARA Method}
\label{fig:laurae_rf_ablation}
\end{figure}

\newtext{We present the results for RSRS implemented with mBERT, XLM-R, and a monolingual BERT-based model in Table~\ref{tab:rsrs_ablation_results}. Although RSRS with the monolingual BERT-based models is the highest performing method on 6 of the 14 datasets, the RSRS variant with mBERT has the highest average performance. Our findings algin with~\citet{naous-etal-2024-readme}, who found that RSRS implemented with multilingual BERT models had better average performance than RSRS implemented with monolingual BERT-based models. Thus, we report RSRS with mBERT as the underlying PLM in the main results. However, even if we were to replace the results in Table~\ref{tab:laurae_results} with another RSRS variant, it would not change our main findings. LAURAE would still be the highest performing method on 13 of 14 datasets. }

\section{LAURAE with RSRS}
\label{sec:laurae_rsrs}

In Figure~\ref{fig:laurae_rf_ablation}, we plot the performance difference between using readability formulas as the shallow unsupervised method in LAURAE and using the RSRS as the shallow unsupervised method. It shows that LAURAE with readability formulas outperforms LAURAE with RSRS on 9 of 14 datasets. The average performance increase is 0.005 points. Although a relatively small average difference, we combine this with the fact that, in comparison to RSRS, readability formulas are less computationally expensive, more interpretable, and have more user-friendly platforms (e.g., Readable.com) that allow less technical users to implement them. Thus, we recommend implementing LAURAE with readability formula scores for the shallow feature score.

\section{Verbal Confidence Score Distribution and LAURAE-agg performance}
\label{app:laurae_agg}

\begin{table}[t!]
\resizebox{\linewidth}{!}{
\begin{tabular}{|>{\centering \arraybackslash}m{0.26\linewidth}|>{\arraybackslash}m{0.22\linewidth}|>{\arraybackslash}m{0.24\linewidth}|>
{\arraybackslash}m{0.14\linewidth}|>
{\arraybackslash}m{0.12\linewidth}|}
\hline

\textbf{Dataset} & \textbf{LAURAE} & \textbf{LAURAE-agg} & \textbf{Mean} & \textbf{Std Dev} \\
\hline
Greek Lang.  & 0.430 & 0.436 & 7.170 & 0.486  \\
\hline
Greek Hist. &0.572 & 0.568 & 7.736 &  0.484 \\
\hline
Vikidia (fr) & 0.953 & 0.960 & 7.563 & 0.439  \\
\hline
Vikidia & 0.900 &  0.900 &  8.025 & 0.447  \\
\hline
Asset &  0.629 & 0.627 & 8.032 & 0.283 \\
\hline
CLEAR & 0.735 & 0.736 & 7.962 & 0.187 \\
\hline
OneStop & 0.654 & 0.654 & 7.934 & 0.076 \\
\hline
MedReadMe & 0.770 & 0.772 &  8.223 & 0.355 \\
\hline
Cambridge & 0.860 & 0.860 & 8.183 & 0.373 \\
\hline
ReadMe & 0.798 & 0.796 & 8.096 & 0.465 \\
\hline
ReadMe (fr) & 0.750 &  0.748& 7.975 &  0.288  \\
\hline
ReadMe (hi) & 0.754 & 0.752 & 7.978 & 0.365  \\
\hline
ReadMe (ar) & 0.757 & 0.760 & 7.905 & 0.433 \\
\hline
ReadMe (ru) & 0.803 & 0.806 &  7.886 & 0.321  \\
\hline
\hline
\textit{Total} & \textit{0.740} & \textit{0.741} & \textit{7.955} & \textit{0.422} \\
\hline
\end{tabular}
}
\caption{\centering Verbal Confidence Score Distribution and Performance Evaluation of proposed LAURAE-agg}
\label{tab:laurae_agg_results}
\end{table}

\begin{figure}[t!]
\includegraphics[width=0.96\linewidth]{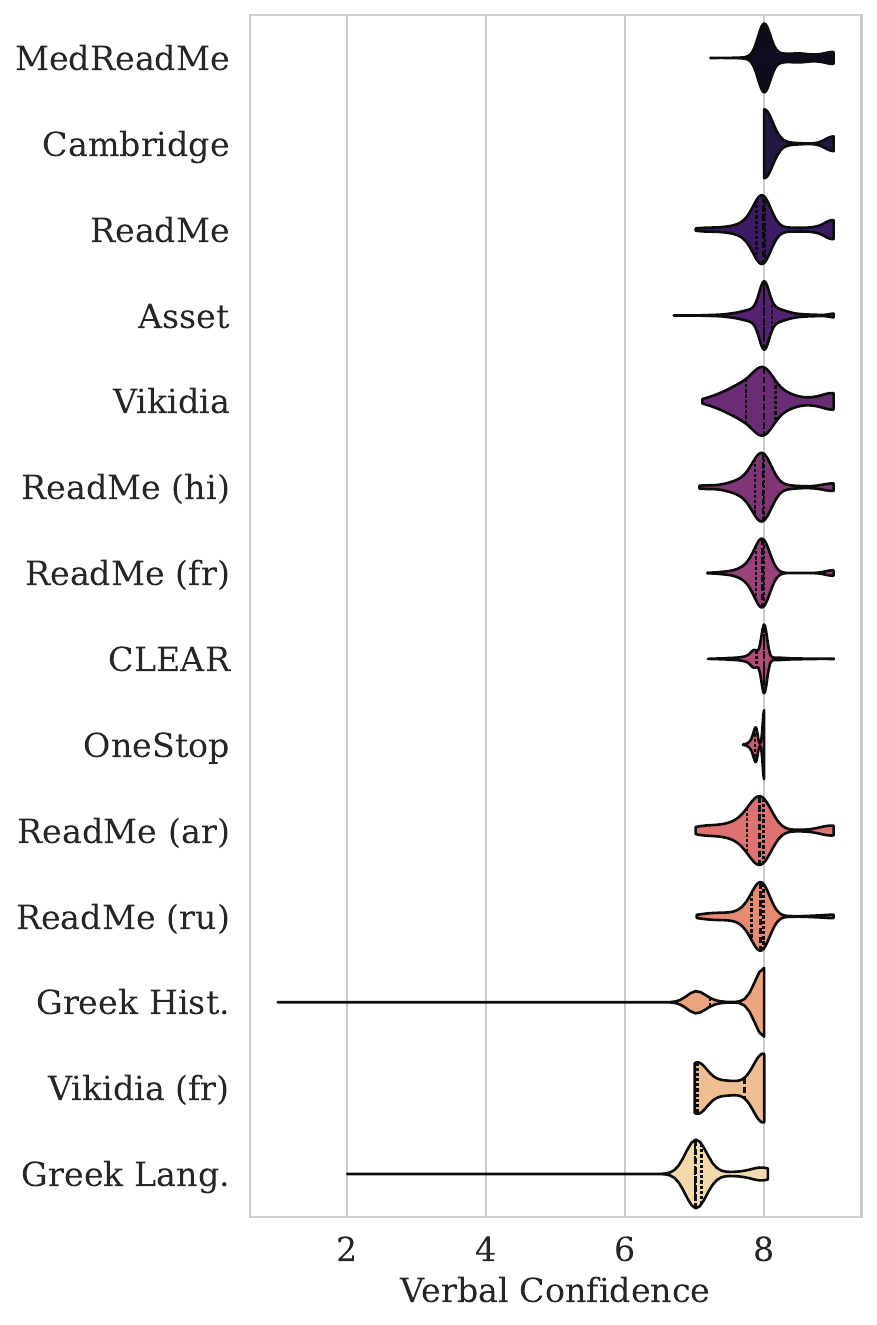}
\caption{\centering Distribution of Verbal Confidence Scores by Dataset}
\label{fig:vconf_distribution}
\end{figure}

The distribution of confidence scores (means/standard deviations reported in Table~\ref{tab:laurae_agg_results}, and distributions visualized in Figure~\ref{fig:vconf_distribution}) raises several areas for future research. 

First, both within and across datasets, verbal confidence scores are clustered near the high end of the range (i.e., 1-9) that LLMs were prompted to provide confidence scores on. Even for datasets where performance is poor (e.g., Greek textbook datasets), the confidence scores are still rarely under 7. Thus, it would be an important area of future research to determine if there are techniques to improve confidence score calibration. This may include adaptations from prior works that prompt LLMs to explicitly consider other answers in their explanations~\cite{tian-etal-2023-just}.

Second, we found that LLM confidence scores are systematically lower when texts are non-English. On average, confidence in non-English datasets is just 0.7678, compared to 0.8071 for English datasets. These differences are not fully explained by lower performance on non-English datasets, nor by the use of Aya 32B model for non-English texts. For example, we compared the verbal confidence and actual performance using the Llama 70B model for the English Vikidia and French Vikidia datasets. While accuracy is higher on the French Vikidia dataset (0.947 versus 0.887), the verbal confidence scores are higher for the English Vikidia dataset  (8.025 versus vs. 7.571). This suggests that LLMs may be relatively under-confident when rating non-English texts (likely due to a smaller amount of training data compared to English texts), and this encourages future research to assess and improve LLM confidence calibration across languages.

\end{document}